\documentclass{sigkddExp}

\usepackage{algorithm}
\usepackage{algpseudocode}
\usepackage{booktabs}
\usepackage{float}

\begin{document}
%

\title{DBR: Divergence-Based Regularization for Debiasing \\ Natural Language Understanding Models}
%

\numberofauthors{1}
%



\author{Zihao Li\textsuperscript{1}\thanks{\, Work done during Zihao Li's remote internship at NJIT.}, Ruixiang Tang\textsuperscript{2}, Lu Cheng\textsuperscript{3}, \
Shuaiqiang Wang\textsuperscript{4}, Dawei Yin\textsuperscript{4}, Mengnan Du\textsuperscript{1}\\
\textsuperscript{1}New Jersey Institute of Technology,
\textsuperscript{2}Rutgers University,
\textsuperscript{3}University of Illinois Chicago,
\textsuperscript{4}Baidu\\
{lizihao9885@gmail.com}, 
{mengnan.du@njit.edu}
}
\date{30 July 1999}
\maketitle
\begin{abstract}
Pre-trained language models (PLMs) have achieved impressive results on various natural language processing tasks. However, recent research has revealed that these models often rely on superficial features and shortcuts instead of developing a genuine understanding of language, especially for natural language understanding (NLU) tasks. Consequently, the models struggle to generalize to out-of-domain data. In this work, we propose Divergence Based Regularization (DBR) to mitigate this shortcut learning behavior. Our method measures the divergence between the output distributions for original examples and examples where shortcut tokens have been masked. This process prevents the model's predictions from being overly influenced by shortcut features or biases. We evaluate our model on three NLU tasks and find that it improves out-of-domain performance with little loss of in-domain accuracy. Our results demonstrate that reducing the reliance on shortcuts and superficial features can enhance the generalization ability of large pre-trained language models.

\end{abstract}

\section{Introduction}
\vspace{3ex} 
Pre-trained language models (PLMs), such as BERT~\cite{devlin2018bert}, RoBERTa~\cite{liu2019roberta}, and Electra~\cite{clark2020electra}, have achieved impressive results on various natural language understanding (NLU) tasks. However, recent studies suggest that these PLMs heavily rely on a phenomenon called ``\underline{shortcut learning}''~\cite{10416838,wang2024towards}, where they capture shallow correlations between labels and shortcut features of examples instead of developing a deeper semantic understanding of language~\cite{clark2019don,du2022shortcut}. In natural language inference, for example, which involves determining the logical relationship between two sentences, recent research indicates that models often associate negative or contradiction labels with specific negation words such as ``no,'' ``none,'' or ``not.'' Due to this shortcut learning, these biased models demonstrate impressive performance for in-domain data by exploiting spurious patterns but struggle to generalize to out-of-domain data.

Correcting these biases and training more robust models has recently attracted significant interest~\cite{du2022shortcut,lyu2023feature,liu2024self}. Most existing debiasing methods relied on some prior human knowledge to identify bias types like partial-input bias~\cite{gururangan2018annotation,poliak2018hypothesis} and lexical overlap~\cite{mccoy2019right}. To address this issue automatically without specifying bias types, efforts have been made to propose debiasing methods that eliminate spurious correlations and improve OOD performance. These approaches include instance reweighting~\cite{schuster2019towards,utama2020towards}, confidence regularization~\cite{utama2020mind}, and product of experts~\cite{clark2019don,sanh2020learning}. 

Despite the recent advancements, effectively addressing bias in NLU models remains a challenging task. There are two primary challenges associated with existing debiasing methods. Firstly, most existing methods rely on training a ``bias-only model" to assist in the debiasing process, which allows the debiased model to focus on specific examples. However, the generalization performance of debiased models heavily depends on humans' prior knowledge about biases in the training data. Unfortunately, this prior knowledge can only identify a limited number of biases in the data. Although it is possible to reduce the use of some known shortcuts, models may still exploit other shortcuts for prediction. This could explain why existing mitigation methods only provide limited gains in generalization~\cite{du2022shortcut}. 
Therefore, we need to reduce the reliance on bias-only models. Secondly, current debiasing methods are often treated as black boxes, since it is unclear how these models actually improve generalization, and whether they genuinely reduce their dependence on superficial features. This lack of transparency hinders the ability to understand the underlying mechanisms of improvement and limits further advancements of pre-trained language models in model generalization.

\begin{figure*}[t]
\begin{center}
\centerline{\includegraphics[width=2\columnwidth]{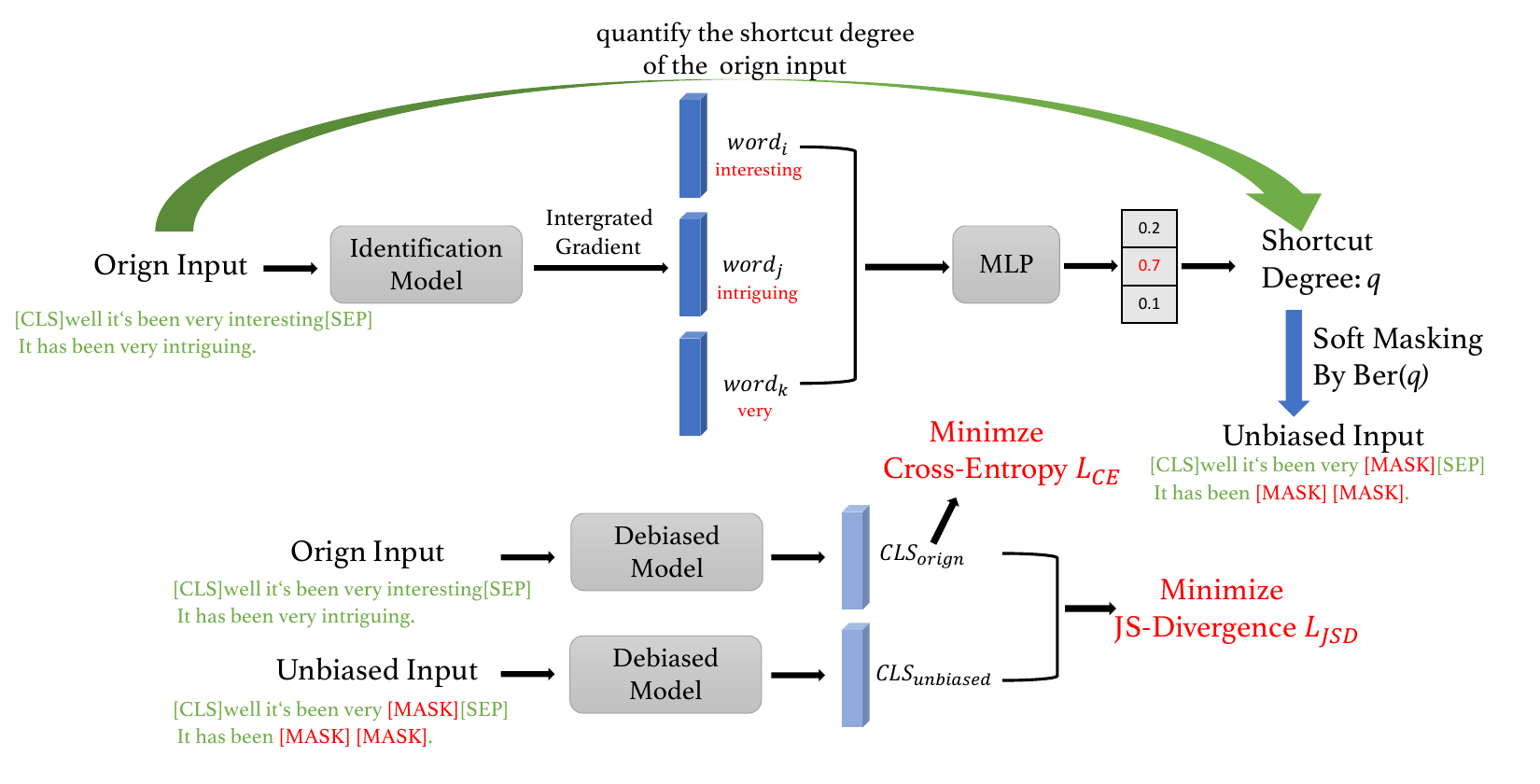}}
    \caption{The proposed DBR framework. We first train a shortcut identification model to compute the shortcut degree of each sample, then use the regularization loss based on the JSD divergence to train the debiased model.}
    \label{fig:main_framework}
    \end{center}
\end{figure*}

To address these research challenges, we propose Divergence Based Regularization (DBR), a \emph{transparent} approach to explicitly enforce the model to reduce reliance on shortcut features (Figure~\ref{fig:main_framework}), thereby improving the robustness of NLU tasks. Specifically, we first mask shortcut tokens to prevent the prediction of the model from being affected by them. In this way, we can construct unbiased versions of original examples, then add a regularization loss to make the original and unbiased examples' representations as close as possible.
However, not all examples exhibit shortcut behavior. Applying this process to all examples would damage semantic meaning. We use a bias-only model to determine which examples actually rely on shortcuts, calculating each example's probability of being a shortcut example. We then use the soft masking strategy of our proposed method to softly mask salient tokens based on these probabilities. This soft masking strategy generates different masked examples for each epoch, improving the model's robustness. 
We evaluate DBR on three common NLU tasks, and the results indicate that our approach improves out-of-domain performance.
The major contributions of this work can be summarized as follows:

\begin{itemize}\setlength\itemsep{-0.3em}  
    \item We propose DBR, a debiasing framework to discourage the NLU models from relying on the shortcut tokens for prediction.
    \item Our proposed DBR method reveals deeper factors that affect model robustness, including the impact of token-level factors. 
    \item Experimental results over three NLU tasks show improved OOD performance, demonstrating that our DBR method reduces shortcut learning and improves generalization.
\end{itemize}

\section{Related Work}
\vspace{3ex}
In this section, we summarize two lines of research that are most relevant to ours.

\noindent\textbf{Data Bias and Shortcut Learning.}\quad
Textual data contain various types of biases, such as word co-occurrence~\cite{gururangan2018annotation}, lexical overlap~\cite{mccoy2019right}, partial inputs~\cite{gururangan2018annotation,poliak2018hypothesis}, and negation words~\cite{utama2020towards}.
Models trained on such biased data will capture spurious correlations in the data without achieving true semantic understanding. This phenomenon is known as \emph{shortcut learning}.
One study models the distribution of shortcut words as a long-tail distribution and uses its characteristics to debias models~\cite{du2021towards}.
Most shortcut phenomena stem from the co-occurrence of specific words and labels. For example, negation words like ``no'' and ``none'' often correlate with contradiction labels in natural language inference tasks~\cite{gururangan2018annotation}.
Recent studies have shown that shortcut learning can negatively impact model performance on OOD datasets~\cite{geirhos2020shortcut,gururangan2018annotation}.

\vspace{2pt}
\noindent \textbf{Shortcut Mitigation.}\quad 
Clark et al. proposed a Product of Experts method that combines a bias-only model's knowledge with a base model~\cite{clark2019don}. It first trains a bias-only model and then uses its predictions to train a robust model \cite{schuster2019towards}.
Similar to focal loss \cite{lin2017focal}, example reweighting \cite{clark2019don} improves models by down-weighting overconfident examples, i.e., shortcut examples.
Confidence regularization \cite{utama2020mind} encourages models to reduce confidence in predictions for biased samples.
Soft label encoding proposed to train a teacher model to determine the shortcut degree, then the degree is used to generate soft labels for robust model training~\cite{he2023mitigating}. DCT employs a positive sampling
strategy to mitigate features in the sample~\cite{lyu2023feature}.

In contrast to these previous methods, our proposed framework takes a more direct approach by explicitly suppressing the NLU model's ability to capture undesirable correlations between shortcut tokens and certain labels. This is achieved through a combination of strategic token masking and distribution alignment, providing a more transparent way to reduce shortcut reliance while maintaining model performance.

\section{Proposed Method}
\vspace{3ex}
In this section, we give a detailed introduction to the proposed Divergence Based Regularization (DBR) debiasing framework  (Figure~\ref{fig:main_framework}). 
It should be noted that, the proposed method is aimed at debiasing PLMs belonging to the traditional pre-training and finetuning paradigm (such as BERT) that are prone to suffer from shortcut learning issue.

\subsection{Proposed Debiasing Scheme}
\vspace{3ex}
The goal of NLU is to classify the semantic relationship between two sentences as one of multiple classes, and we formulate it into a multi-class classification task. Given a pair of a sentence $x_i \in \mathcal{X}$  and its label $y_i  \in \mathcal{Y}$, we aim to learn a robust mapping function of $\mathcal{F}$ $:x_i \rightarrow y_i$. We follow the standard pre-training and fine-tuning paradigm. The model should rely on semantic understanding for prediction rather than relying on shortcuts, so that it can generalize well to out-of-domain datasets.  

\vspace{2pt}
The key motivation of our approach is to discourage excessive reliance of NLU models on shortcuts. We propose to achieve this
by masking shortcut tokens and aligning the prediction distributions between the original and masked samples. Specifically,
our framework consists of two stages. We first develop a shortcut identification model using the training data to detect linguistic shortcuts in the text (Section \ref{sec:Shortcut-Features-Identification}). Subsequently, in the second stage, we train a debiased model by introducing a regularization loss that focuses on aligning distributions (Section \ref{sec: hard masking}). 
More specifically, during the second stage, we mask the shortcut tokens and encourage the NLU model to generate similar prediction 
distributions for both the original samples and the samples with masked shortcut tokens. To ensure that the semantic meaning of the text remains unaffected, we employ the soft masking strategy to further refine the masking process (Section \ref{sec: soft masking}).  

\subsection{Shortcut Tokens Identification}\label{sec:Shortcut-Features-Identification}
\vspace{3ex}
To effectively capture shortcut features in the sample and analyze the factors influencing model robustness in detail, we utilize a gradient-based interpretation technique known as Integrated Gradients (IG) \cite{sundararajan2017axiomatic}. This method enables us to determine the impact of each token on the model's prediction, aligning perfectly with our requirements.
By attributing the ground-truth label to each input token, IG generates interpretations for every token in the text. The outcome is presented as a feature importance vector, indicating the significance of each token.
The main steps of IG are described as follows. We first construct a baseline input $x_{base}$ with the same dimensions as the original input $x_i$, and then integrate the gradients of prediction probability w.r.t. $m$ intermediate samples from the baseline input $x_{base}$ to the original input $x_i$. It can be formulated as follows:
$\scriptstyle$\begin{equation}
    g_{x_i} = (x_i - x_{base})\cdot \frac{1}{m}\sum\limits_{k=1}^{m}\nabla_{x_i} f_{y}\big(x_{base} + \frac{k}{m} (x_i-x_{base})\big)\label{IG}.
\end{equation}

The shape of  the original input $x_i$ 
is $(L, d)$ with $L$ tokens, and each token represents its word embedding with $d$ dimensions. We employ all-zero word embeddings to represent the baseline input $x_{base}$. As such, we obtain $g_{x_i}$, i.e., the attribute vector for each token, with the same shape as $x_i$. To compute the attribution of each token, we compute the $\ell_2$ norm of each attribution vector to measure the attribution of each token. Shortcut words mean that the prediction  highly relies on these words, thus the shortcut words can be regarded as tokens with high attribution to prediction. So we select top-$N$ tokens  by their attribution values as our shortcut tokens of the input text $x_i$.

\subsection{Debiasing by Hard Masking}\label{sec: hard masking} 
\vspace{3ex}
In this section, we introduce the details of the divergence based regularization for debiasing NLU models.
We first (hard) mask of the shortcut tokens identified within the original sentence, to acquire an unbiased representation of the original sample, denoted as $x_{unbiased}$. By masking these tokens, which significantly influence the model's predictions, we ensure that the model is not influenced by these shortcuts when making predictions. Consequently, the sample with the masked shortcut words can be considered an approximately unbiased representation of the original sample.

Inspired by \cite{guo2022auto}, after obtaining the unbiased representation of the sample, we align the distribution space of the unbiased sample $x_{unbiased}$ with that of the original sample $x_{original}$. This helps mitigate the influence of shortcut tokens on the model. We use the Jensen-Shannon divergence (JSD) \cite{fuglede2004jensen}, a function for measuring the distance between probability distributions as our regularization loss function to minimize the disagreement between the distributions of the unbiased sample and the original sample. Compared with the Kullback–Leibler divergence (KLD) loss, the JSD loss is a symmetric representation of the latter. It can be described as:
\begin{equation}
\begin{aligned}
    & JSD(p_{1},p_{2}) = \frac{1}{2}\sum\limits_{i=1}^{2}(KLD(p_{i}||\frac{p_1+p_2}{2}),\\
    & KLD(p_1||p_2) = \sum_{k\in \mathcal{Y}}p_{1}(k)log(\frac{p_{1}(k)}{p_{2}(k)}).
\end{aligned}
\end{equation}

We compute the JSD score between the distribution of unbiased sample $p_{unbias}=p([CLS] = y_i|\mathcal{F},x_{unbias})$ and that of the original sample $p_{orign} = p([CLS] = y_i|\mathcal{F}, x_{orign})$. Our goal is to minimize the JSD score between $p_{unbias}$ and $p_{orign}$ to discourage the model from relying on shortcut tokens for prediction.

\subsection{Debiasing by Soft Masking}\label{sec: soft masking}
\vspace{3ex}
The aforementioned hard masking scheme has two limitations. Firstly, the top-$N$ shortcut tokens selected in Section~\ref{sec: hard masking} may not accurately represent the actual shortcut tokens. There is a possibility that tokens that positively contribute to the model's prediction are genuine important tokens. Secondly, the hard masking scheme, which masks the text input, can potentially impact the semantic meaning of the text. For instance, in the sentence ``\emph{The movie I saw last night is so excellent}," the hard masking scheme might mask the word ``excellent," which significantly contributes to the predicted label and also conveys important semantic information. Recent research~\cite{wang2021identifying,srivastava2020robustness,wang2020identifying} has shown that ``genuine'' tokens typically play a vital role in conveying semantic meaning, and their correlation with labels is what the model aims to capture. On the other hand, the correlation between ``spurious'' tokens (i.e., non-genuine shortcut tokens) and labels fails to generalize to OOD datasets. Therefore, while masking shortcut tokens can enhance the model's generalization, masking genuine tokens can compromise the semantic meaning of the text and hinder the model from capturing the relationship between the text and the label. These two limitations of hard masking motivate the design of the \emph{soft masking} strategy.

\vspace{2pt}
\noindent\textbf{Quantifying Shortcut Degree.}\quad
The first step is to quantify the shortcut degree of each training sample.
After obtaining the shortcut tokens for each sample, we design a metric to measure the amount of shortcut information contained in each sample. Based on a simple intuition, if a sample contains a significant amount of shortcut information, the model can easily predict the label solely based on these shortcut words. Motivated by this idea, we train a bias-only model that takes shortcut words as input and predicts the label. Consequently, we obtain the prediction probabilities for each label. Now, the question is how to quantify the shortcut degree based on these probabilities. It is widely accepted that the more biased the model is, the more confident it becomes in its predictions, resulting in higher variance in the predicted outcomes~\cite{huangbiaspad}. Therefore, we utilize the variance (Var) metric to represent the shortcut degree of a sample $i$, which is described as follows:
\begin{equation}
    \hat{s_i}^2 =  \frac{s_{i}^2-min\{s_{j}^2\}_{j=1}^{m}}{max\{s_{j}^2\}_{j=1}^{m}-min\{s_{j}^2\}_{j=1}^{m}}, s_{i}^2 = \frac{\sum\limits_{j=1}^{K}(p_j - \overline{p})^2}{K-1},\label{var}
\end{equation}
where $K$ denotes the number of labels, $p_{j}$ denotes the probability of predicting label $j$, and $m$ denotes the batch size. The example with high variance can be considered as an overconfident or shortcut example~\cite{huangbiaspad}. We normalize the sample variance $s_{i}^2$ to $\hat{s_{i}}^2$ within the same batch to make it range from 0 to 1. $\hat{s_{i}}^2$ can be considered as shortcut degree of sample $i$.

\vspace{2pt}
\noindent\textbf{Soft Masking Framework.}\quad
Once we have determined the shortcut degree of each training sample, we proceed to implement a soft masking strategy. In this strategy, the decision of whether to mask the top-$N$ shortcut tokens in a sample is made using the Bernoulli distribution. This distribution determines whether each token should be masked or left unchanged based on a probability threshold. The soft masking can be described as follows:
\begin{equation}
    x_{unbias} = \mathcal{M}(x_i),\mathcal{M} \sim Ber(\hat{s_{i}}^2)\label{ber},
\end{equation}
where $\mathcal{M}$ denotes the operation of masking shortcut tokens of $x_i$, $\hat{s_{i}}^2$ denotes the normalized variance, representing the probability that the sample belongs to the shortcut sample. 
Therefore, the final loss can be defined as follows:
\begin{equation}
    \mathcal{L} = \mathcal{L}_{CE}(\mathcal{F}(x_i),y_i) + \lambda \mathcal{L}_{JSD}(p_{unbias},p_{orign}),
    \label{equ:loss}
\end{equation}
where $\mathcal{L}_{CE}$ denotes the cross-entropy loss and $\lambda$ is the weight of $JSD$ loss function.
The overall steps of DBR are given in Algorithm 1.

\setlength{\textfloatsep}{14pt}

\begin{algorithm}[t]
\caption{Pseudo-code for DBR framework}
\begin{algorithmic}[1]
\State \textbf{Input:} Training data $\mathcal{D} = \{(x_i, y_i)\}_{i=1}^N$, Identification model $\mathcal{F}_i$, Bias-only model $\mathcal{F}_{\text{bias}}$.
\State \textbf{Output:} Debiased model $\mathcal{F}_{\text{debias}}$.

\State //obtain top-$N$ tokens list $S_i$ using identification model $\mathcal{F}_i$ for each sample $x_i\in \mathcal{D}$  by Equation~\ref{IG};
\For{$(x_i, y_i) \in \mathcal{D}$}
    \State $S_i$ = \{\};  
    \State $S_i$ $\leftarrow$ top-$N_{x_{i}}$ tokens by $g_{x_{i}}$;
    \State //where $g_{x_{i}}$ obtained from Equation~\ref{IG}
\EndFor

\State  // Get the shortcut degree $\hat{s_i}^2$ for each sample $x_i\in \mathcal{D}$
\For{$(x_i, y_i) \in \mathcal{D}$}
    \State $p$ = $\mathcal{F}_{bias}(S_i)$
    \State // obtain $\hat{s_i}^2$ by Equation~\ref{var}
\EndFor

\State // Train the debiased model.
\For{$(x_i, y_i) \in \mathcal{D}$}
    \State// Using shortcut degree $\hat{s_i}^2$ obtained from above to generate unbias sample by Equation~\ref{ber}
    \State Training the model using the loss function $\mathcal{L} = \mathcal{L}_{CE}(\mathcal{F}_{debias}(x_i),y_i) + \lambda \mathcal{L}_{JSD}(p_{unbias},p_{orign})$
    
\EndFor

\end{algorithmic}
\label{alg:mask_algorithm}
\end{algorithm}

\begin{table*}
\centering
\scalebox{1.0}{
\begin{tabular}{lcccccccc}
\toprule
{\textbf{Method}} & \multicolumn{3}{c}{\textbf{MNLI}} & \multicolumn{3}{c}{\textbf{FEVER}}& \multicolumn{2}{c}{\textbf{QQP}}\\
\cmidrule(lr){2-4}\cmidrule(lr){5-7}\cmidrule(lr){8-9} & dev & HANS & dev hard &  dev &Sym1&Sym2 & dev & PAWS\\
\hline
BERT-base & 84.5 & 62.4& 77.0 & 85.6 &55.1& 63.1 & 91.0 & 33.5\\
\hline

ER\cite{schuster2019towards}&81.4&68.6&-&87.2&-&65.6&85.2&57.4\\
PoE\cite{he2019unlearn,clark2019don}&80.7&68.5&-&85.4&-&65.3&-&-\\
ConRe\cite{utama-etal-2020-mind}&83.9&67.7&-&\textbf{87.9}&-&66.1&89.0&43.0\\
Learned-Mixin\cite{clark2019don}&84.3&64.0&-&83.3&60.4&64.9&86.6&\textbf{56.8}\\
Modeling Bias\cite{mahabadi2019end}&84.2&64.7&76.8&86.5&-&66.3&-&-\\
Soft Label\cite{he2023mitigating}&81.2&68.1&-&87.5&60.3&66.9&-&-\\
Debias Mask\cite{meissner2022debiasing}&81.8&68.7&-&84.6&-&64.9&-&-\\
DCT\cite{lyu2023feature}&84.2&68.3&-&87.1&\textbf{63.3}&\textbf{68.4}&-&-\\
\hline

DBR-soft mask & \textbf{84.5} & \textbf{68.6} & \textbf{78.8} & 86.4 & 59.2&66.4&\textbf{90.7}&41.8\\
DBR-hard mask& 83.9 & 67.4 & 78.0 &85.4&55.1&64.9&90.3&41.2\\
\bottomrule
\hline
\end{tabular}}
\caption{Performance between DBR and other baselines on three NLU tasks. For MNLI task, we choose dev-mismatch as our dev set. The results for the baselines of ER, POE, ConRe are taken from [31]. Bold results indicate the best results of the above baseline, excluding BERT-base.}
\label{tab:overall-results}
\end{table*}

\section{Experiments}
\vspace{3ex}
We conduct experiments to evaluate the debiasing performance of the proposed DBR debiasing framework, to answer the following three research questions:

1) In comparison to established baselines, does the proposed DBR debiasing method effectively optimize the trade-off between in-domain and OOD performance? (Section~\ref{sec:OOD-in-domain-tradeoff})

2) Does the proposed soft masking technique prove to be effective in debiasing shortcut learning? (Section~\ref{sec:ablation-study})

3) What are the key factors that contribute to the effectiveness of the proposed method? (Section~\ref{sec:detailed-analysis})

\subsection{Experiment Settings}
\vspace{10pt}
In this section, we present the comprehensive experimental setup used to evaluate our proposed DBR framework. We describe the datasets used for three NLU tasks, baseline methods for comparison, and implementation details including model architectures and training configurations.

\subsubsection{Datasets}
\vspace{3ex}
We evaluate the generalization performance of DBR in three common NLU tasks.

\noindent \textbf{Nature Language Inference.}\, The training dataset of this task is MNLI~\cite{williams2017broad} which contains 392,702 samples. Each training sample consists of two sentences representing the premise and the hypothesis, the goal of the task is to predict whether the relationship between the premise and the hypothesis is entailed, contradicted, or neutral. There are two development sets of MNLI: dev-matched containing 9,815 samples and dev-mismatched 9,832 samples. The difference between them is that dev-matched is consistent with the source of the training datasets and dev-mismatched is not. 
For OOD test sets, we employ HANS~\cite{mccoy2019right} and MNLI-hard~\cite{gururangan2018annotation} for evaluation. 

\vspace{2pt}
\noindent \textbf{Fact Verification.}\,
FEVER~\cite{thorne2018fever}, which comprises 242,911 samples, is the training set. Each training sample consists of two sentences, representing the claim and the evidence. The objective of the task is to predict the relationship between the claim and the evidence, categorizing it as ``refute," ``support," or ``not enough information." Additionally, we have a development set containing 16,664 samples, which will be used for evaluating and fine-tuning our model. Symmetric v1 and v2 (Sym 1 and Sym 2)~\cite{schuster2019towards} are the OOD test sets. Both test sets are synthesized and specifically created by introducing perturbations to the sentence pairs. These perturbations are designed to challenge the model and result in poor performance if no debiasing strategies are applied. Each synthesized test set include 712 samples.

\noindent \textbf{Paraphrase Identification.}\,
For QQP dataset, the task aims at predicting whether the relationship between the sentence pair is duplicate or not. The training set contains 363,846 samples and the development set contains 404,30 samples. 
We use Quora Question Pairs (QQP)\footnote{ https://www.kaggle.com/c/quora-question-pairs} as our training dataset, and use PAWS
as our challenging OOD test set \cite{zhang2019paws}. The adversarial samples are generated with lexical-overlap bias.

\begin{figure*}[ht]
\begin{center}
\centerline{\includegraphics[width=2.1\columnwidth]{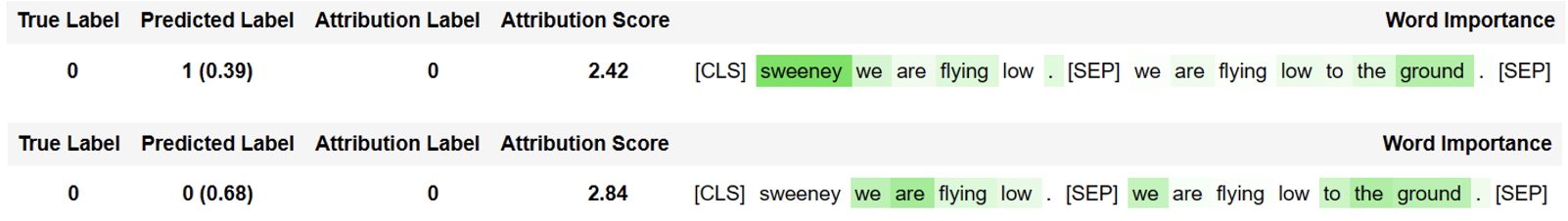}}
    \caption{Attribution result visualization,  the first and second row denote the attribution of each word before mitigation and after mitigation respectively. Words marked in green represent that the word contributes to the model prediction results, and the darker the color, the greater the contribution.}
    \end{center}
    \label{fig:vis}
\end{figure*}

\subsubsection{Comparing Baselines}
\vspace{3ex}
We compare the proposed DBR method, including soft mask and hard mask versions, with several representative debiasing baselines, detailed as follows. 

\noindent \textbf{Example Reweight (ER)}~\cite{schuster2019towards}\, ER first trains a basic model to obtain predictions with bias, then trains a debiased model using the following loss:
$
    \mathcal{L} = -(1-p_{b}^{i})y^{i} \cdot p_{d}^{i},
$
where $p_b$ and $p_d$ denote the softmax output of the basic model and debiased model, respectively.
Examples with high confidence are allocated with less attention.

\noindent \textbf{Product of Experts (PoE)}~\cite{he2019unlearn,clark2019don}\, PoE first trains a basic model and combines the softmax output of it and the debiased model. The ensemble loss is described as:
$
    \mathcal{L} = - y^{i} \cdot \log softmax(p_{b}^{i} + p_{d}^{i}).
$

\noindent \textbf{Confidence Regularization (ConRe)}~\cite{utama-etal-2020-mind}\, ConRe encourages the student model to assign less attention to samples that the teacher model considers biased:
$
    \mathcal{L} = -S(p_t, p_{b}^{i}) \cdot \log p_d,
$
where
$S(p_t, p_{b}^{i})$ denotes the soft predictions  with temperature $p_{b}^{i}$.

In addition to above basic debiasing methods, we also compare our method with some complex baselines such as {Learned-Mixin}\cite{clark2019don}, {Modeling Bias}~\cite{mahabadi2019end}, {Soft Label Encoding}~\cite{he2023mitigating}, {Debias Mask}~\cite{meissner2022debiasing}, and {DCT}~\cite{lyu2023feature}.

\subsubsection{Implementation Details}
\vspace{3ex}
In our experiments, we utilize the BERT-base-uncased model\footnote{https://huggingface.co/bert-base-uncased} as the backbone for both the identification model and the debiased model. This model consists of 12 Transformer blocks, each with a hidden layer dimension of 768. For the bias-only model, we adopt a simple structure to effectively identify shortcut samples. It consists of a single Multi-Layer Perceptron (MLP) with the Rectified Linear Unit (ReLU) activation function. More detailed information about the bias-only model can be found in Section \ref{sec:learning of bias-only}.

Regarding the training settings, we train the identification model and the debiased model using the entire training dataset for 12 epochs. In contrast, the bias-only model is trained using a smaller subset of the training dataset for only 1 epoch to mitigate the risk of overfitting. The batch size for training is set to 32 for the identification model and 18 for both the bias-only model and the debiased model. The learning rate is set to $2e-5$. During the ``top-$N$" selection process, the value of $N$ is set to 3. The maximum length of the input sequence is limited to 512 tokens. The hidden layer dimension of the MLP in the bias-only model is set to 100. In Equation~\ref{equ:loss}, the value of $\lambda$ is set to 1.5 for MNLI/QQP, and 3 for FEVER, respectively.

\subsection{Trade-off between In-domain and OOD}\label{sec:OOD-in-domain-tradeoff}
\vspace{3ex}

The results of DBR and the baselines are presented in Table \ref{tab:overall-results}. We can observe that DBR consistently outperforms BERT-base models on all OOD test sets. Compared to the performance of other methods on most OOD test sets, DBR achieves comparable results without a significant drop in performance on the in-domain test set. This suggests that our method successfully achieves a trade-off between the OOD test sets and the in-domain test set. Specifically, DBR outperforms all the debiasing techniques for HANS and MNLI dev hard.

Notably, DBR achieves similar in-domain performance to BERT-base models on MNLI and QQP and even improves the in-domain performance on FEVER. These results show positive evidence that the proposed soft masking strategy enhances the semantic expression of sentences and reduces the reliance of NLU models on shortcut tokens.
In the case of the QQP dataset, some baseline approaches, such as Learned-Mixin~\cite{clark2019don}, achieve promising performance on the PAWS test set. However, these methods suffer from a significant drop in performance on the in-domain dataset. In contrast, our proposed DBR strikes a balance between the in-domain test set and the OOD test set, achieving competitive performance in both scenarios.

\subsection{Ablation Studies}\label{sec:ablation-study}
\vspace{3ex}
We also present the comparison results between using the soft masking strategy and not using it in Table \ref{tab:overall-results}. From the results, when considering the strategy of masking all sentences (hard mask), we observe a decline in performance on the in-domain test set compared to BERT-base. However, when employing the soft mask strategy, we observe improvements in performance on the in-domain test set. Additionally, on the FEVER dataset, there are further improvements compared to BERT-base. These results show that the soft masking not only helps achieve a better understanding of the sentence's semantics compared to the hard masking but also enhances the overall applicability of our method.

\section{A Closer Look}\label{sec:detailed-analysis}
\vspace{3ex}
In this section, we provide further analysis and discussion of the proposed debiasing algorithm.

\subsection{Generalization Visualization}
\vspace{3ex}

We further conduct a visualization analysis through case studies, as depicted in Figure 2. Prior to mitigation, it is evident that the model predominantly focuses on the shortcut word ``sweeney," as indicated by the strong attention weight assigned to it. However, after applying the mitigation strategy, we observe a notable change in the visualization. In the post-mitigation scenario, we can observe that a greater number of words are highlighted in green compared to the pre-mitigation stage. Furthermore, the color distribution is more uniform, indicating a more balanced contribution from multiple words in the text input. This observation suggests that the model now pays attention to a wider range of words in the input text. Consequently, the model's reliance on the shortcut word is reduced, enabling it to better grasp the semantic meaning of the text. This visualization analysis provides evidence of how DBR debiases the model, by affecting the internal factors of the sentence, thereby enhancing the transparency of the debiasing process.

\subsection{Learning of Bias-only Model}\label{sec:learning of bias-only}
\vspace{3ex}
In this section, we analyze the performance of the bias-only model, i.e., the MLP model shown in Figure~\ref{fig:main_framework}.

\begin{table}
\centering
\renewcommand\arraystretch{1}
\resizebox{0.3\textwidth}{!}{
\begin{tabular}{ccc}
\toprule
\textbf{Datasets}&\textbf{Accuracy}&\textbf{Samples}\\
\hline
MNLI& 97.25&2000\\
FEVER& 95.68&3000\\
\bottomrule
\hline
\end{tabular}}
\caption{Accuracy and training samples of the bias-only model in in-domain test set sample from training set.}
\end{table}

\noindent \textbf{The structure of Bias-only Model.}\, The bias-only model in our design serves the purpose of quantifying the degree of shortcuts in the training samples. Therefore, it should achieve high training accuracy w.r.t. in-distribution training data when using a sufficiently simple model structure, a relatively small training dataset, and a training input that is intentionally biased towards shortcuts. To fulfill these requirements, we construct the model input vector by concatenating the encoded text representation of the top-$N$ shortcut words from each sample. The resulting input shape is ($m$, $N \times \text{dim}$), where $m$ denotes the batch size and $\text{dim}$ represents the hidden size of the BERT encoder. The model architecture consists of only one layer of MLP with the ReLU activation function.

For training, we randomly select 1000, 2000, and 3000 samples from the original training set, respectively, and train the bias-only model for one epoch. The remaining 10000 samples are used as the test set (\emph{both the training and test sets are derived from the MNLI training set}). The accuracy results are presented in Table 2, indicating that the bias-only model achieves high accuracy on the in-distribution dataset by relying solely on a small number of words in the text.
In comparison to previous approaches, our bias-only model exhibits a higher degree of bias, aligning with our objectives. These findings validate that our bias-only model effectively quantifies the degree of shortcuts in the training samples, as it achieves remarkable accuracy on the in-distribution dataset by leveraging only a limited set of words.

\noindent\textbf{What Bias-only Model Learns.}\,We also investigate the distribution of top-$N$ shortcut words. We use local mutual information (LMI)~\cite{schuster2019towards} to measure the correlation between top-$N$ words $\omega$ and labels $l$:
\begin{equation}
    LMI(\omega, l) = p(\omega, l) \cdot log(\frac{p(l|\omega)}{p(l)}),
\end{equation}
where $p(\omega, l) = \frac{count(\omega,l)}{|D|}$,$p(l|\omega) = \frac{count(\omega,l)}{count(\omega)}$, $p(l) = \frac{count(l)}{|D|}$ and $|D|$ is the number of top-$N$ shortcut words of the training set.

\begin{table}
\centering
\resizebox{0.42\textwidth}{!}{
\begin{tabular}{cccc}
\toprule
\multicolumn{2}{c}{\textbf{Entailment}}&\multicolumn{2}{c}{\textbf{Contradiction}}\\
\cmidrule(lr){1-2}\cmidrule(lr){2-4}\textbf{Words}&\textbf{LMI $\times 10^{-3}$}&\textbf{Words}&\textbf{LMI$\times 10^{-3}$}\\

\hline
the&8.88&not&22.7\\
and&2.65&no&22.3\\
can&1.25&never&11.7\\
many&0.93&don&3.87\\
good&0.89&didn&2.63\\
great&0.58&cannot&1.23\\

\bottomrule
\hline
\end{tabular}}
\caption{The LMI of top-$N$ shortcut words in the training set of MNLI with respect to the label of ``entailment" and ``contradiction".}
\end{table}

\begin{figure*}
  \centering
  \includegraphics[width=0.98\linewidth]{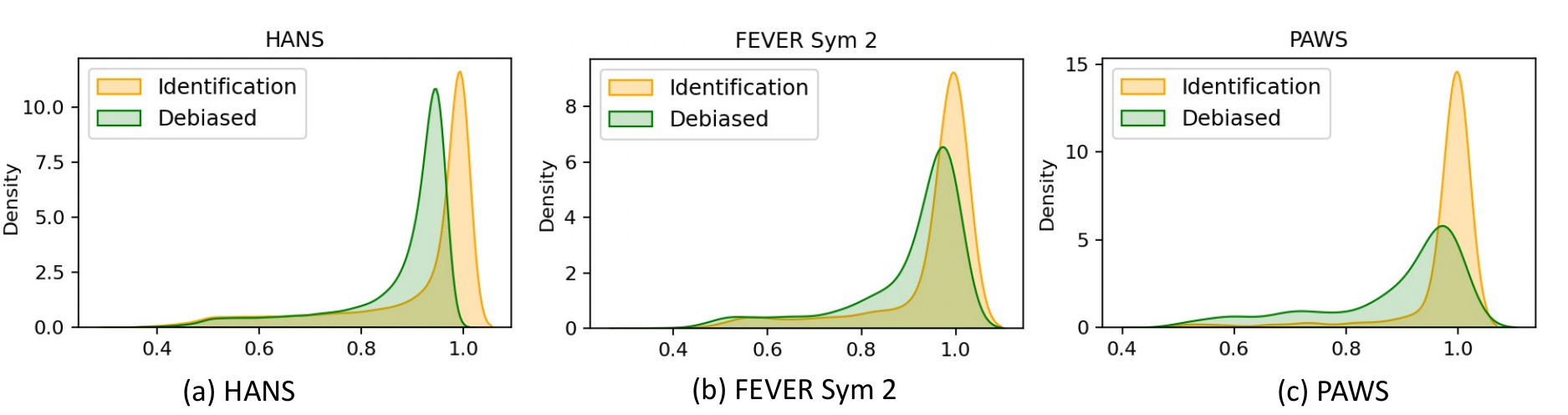}
  \vspace{-7pt}
  \caption{Confidence distribution of the identification model and the debiased model. The orange denotes the identification model and the green denotes the debiased model.}
  \label{fig:overall}
\end{figure*}

In Table 3, we present the selected shortcut words with high LMI that are correlated with the ``entailment" and ``contradiction" labels in the MNLI dataset. An observation from the table is that the majority of words associated with the ``contradiction" label exhibit negative emotions, such as ``no" and ``never," which are highly consistent with the nature of the ``contradiction" label. The same pattern holds true for the words related to the ``entailment" label. Furthermore, we notice that the LMI values for the ``contradiction" label are significantly higher than those for the ``entailment" label. Consequently, to enhance the model's robustness, it becomes crucial to focus on the input associated with negative labels.

Besides, we can find that some words associated with "entailment" can't carry meaningful information regarding the label. Therefore, we filter out the intersection of the top 10 words of the two labels which are considered as words without useful information. We filtered out these high-frequency words and conducted another experiment based on the original settings. The results show that the model performs better in OOD datasets like HANS and FEVER Sym1, and remains the same performance in the original dataset. It shows that masking these high-frequency words that carry little information about the sentence will hinder the model's understanding of the overall semantics.


\begin{table}
\centering
\resizebox{0.4\textwidth}{!}{
\begin{tabular}{l|ccc|ccc}
\hline
& \multicolumn{3}{c|}{MNLI} & \multicolumn{3}{c}{FEVER} \\
Method & dev & HANS & dev hard & dev & Sym1 & Sym2 \\
\hline
DBR-soft mask(filtered) & 84.5 & 68.9 & 78.8 & 86.4 & 59.4 & 66.2 \\
DBR-soft mask & 84.5 & 68.6 & 78.8 & 86.4 & 59.2 & 66.4 \\
DBR-hard mask & 83.9 & 67.4 & 78.0 & 85.4 & 55.1 & 64.9 \\
\hline
\end{tabular}}
\caption{Performance of DBR-soft mask, DBR-hard mask and DBR-soft mask(filtered).}
\end{table}

\subsection{The Convergence of Loss}
\vspace{3ex}
\begin{figure}[t]
\begin{center}
\centerline{\includegraphics[width=0.9\columnwidth]{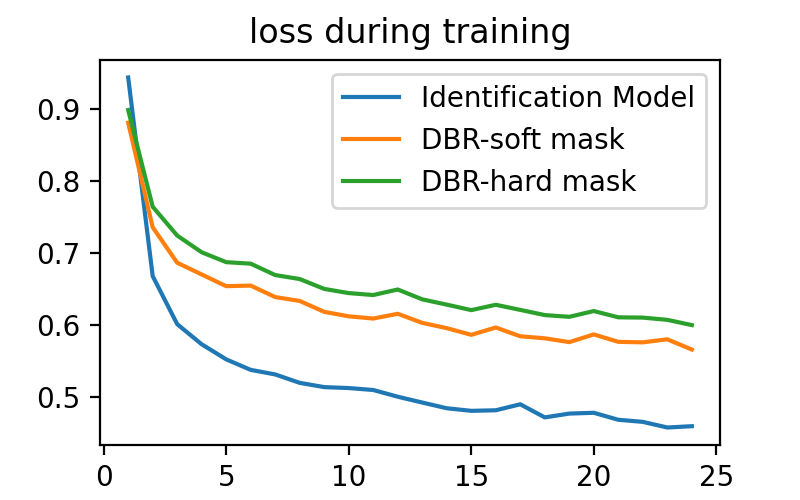}}
    \caption{Loss function curves for three training approaches during the training stage: standard training, DBR-hard mask and DBR-soft mask.}
    \end{center}
    \label{fig:loss_curve}
\end{figure}
In Figure 4, we present the convergence of the loss function during the first epoch for the original training approach and our proposed method. We observe that the loss function of the original training approach converges faster compared to our proposed method. This phenomenon can be explained by the model's tendency to prioritize learning the features of shortcut samples~\cite{du2021towards}. 

The slower convergence of DBR indicates two key points. Firstly, our method focuses more on hard samples rather than shortcut samples, which requires additional training iterations to achieve convergence. Secondly, the slower convergence suggests that DBR effectively guides the model to pay less attention to shortcut features.

\subsection{Confidence Distribution}
\vspace{3ex}
We conduct a comparative analysis of confidence distributions between the identification model and the debiased model, with results presented in Figure 3. A notable observation is that the confidence density curve of the debiased model (green color) shows a leftward shift compared to the identification model (orange color), indicating that DBR successfully reduces overall prediction confidence levels.

This pattern is particularly pronounced in the FEVER and PAWS datasets, where the identification model's curve (orange color) exhibits a steeper profile than the debiased model's curve (green color). This steeper distribution suggests that the identification model produces more concentrated confidence scores. This observation aligns with established findings that models tend to display overconfidence when encountering biased or shortcut examples. Such overconfidence typically manifests in shortcut learning, where models exploit superficial patterns rather than developing deeper understanding.

\section{Conclusions and Future Work} %
\vspace{3ex}
In this work, we have introduced DBR, a novel debiasing approach for natural language understanding models. Our method operates by masking salient words to construct unbiased example representations, then employing a regularization loss to align the distributions between original and unbiased examples. The results show that DBR not only achieves significant improvements in out-of-domain performance but also maintains strong in-domain accuracy.

Moving forward, we plan to explore alternative masking strategies, such as substituting the masked shortcut tokens with alternative tokens.
and extend the debiasing for large language models (LLMs) belonging to the prompting paradigm such as Llama-2~\cite{touvron2023llama}, Mistral~\cite{jiang2024mixtral}.

\newpage

\bibliographystyle{plain}
\bibliography{ref} 
\end{document}